\documentclass[10pt,journal,compsoc]{IEEEtran}
\ifCLASSOPTIONcompsoc
  \usepackage[nocompress]{cite}
\else
  \usepackage{cite}
\fi
\ifCLASSINFOpdf
   \usepackage[pdftex]{graphicx}
\else
\fi
\usepackage{amsmath}
\usepackage{algorithmic}

\usepackage{subfig}
\usepackage{color,soul}
\usepackage{url}
\usepackage[]{footmisc}
\usepackage{ragged2e}

\begin{document}
\title{Bare Demo of IEEEtran.cls for\\ IEEE Computer Society Journals}

\title{A Hamiltonian Monte Carlo Model for Imputation and Augmentation of Healthcare Data}


\author{Narges~Pourshahrokhi,~\IEEEmembership{Student Member,~IEEE,}
Samaneh~Kouchaki,~Kord M. Kober,~Christine~Miaskowski
and~Payam~Barnaghi,~\IEEEmembership{Senior Member,~IEEE,}% <-this % stops a space
\IEEEcompsocitemizethanks{\IEEEcompsocthanksitem 
NP and SK are with CVSSP, University of Surrey, UK.
Email: n.pourshahrokhi@surrey.ac.uk, samaneh.kouchaki@surrey.ac.uk
\IEEEcompsocthanksitem PB is with Department of Brain Sciences Imperial College London and UK Dementia Research Institute. Email: p.barnaghi@imperial.ac.uk
%\IEEEcompsocthanksitem KK is with University of California, San Francisco, US
\IEEEcompsocthanksitem  CM and KK are with Department of Physiological Nursing University of California San Francisco. Email: chris.miaskowski@ucsf.edu, Kord.Kober@ucsf.edu
\IEEEcompsocthanksitem SK and PB are also with the Care Research and Technology Centre at the UK Dementia Research Institute }% <-this % stops an unwanted space
\thanks{Manuscript received .}}

\IEEEtitleabstractindextext{%
\begin{abstract}
\justifying
Missing values exist in nearly all clinical studies because data for a variable or question are not collected or not available. Inadequate handling of missing values can lead to biased results and loss of statistical power in analysis. Existing models usually do not consider privacy concerns or do not utilise the inherent correlations across multiple features to impute the missing values. In healthcare applications, we are usually confronted with high dimensional and sometimes small sample size datasets that need more effective augmentation or imputation techniques. Besides, imputation and augmentation processes are traditionally conducted individually. However, imputing missing values and augmenting data can significantly improve generalisation and avoid bias in machine learning models. A Bayesian approach to impute missing values and creating augmented samples in high dimensional healthcare data is proposed in this work. We propose folded Hamiltonian Monte Carlo (F-HMC) with Bayesian inference as a more practical approach to process the cross-dimensional relations by applying a random walk and Hamiltonian dynamics to adapt posterior distribution and generate large-scale samples. The proposed method is applied to a cancer symptom assessment dataset and confirmed to enrich the quality of data in precision, accuracy, recall, F1 score, and propensity metric.
\end{abstract}

}

\maketitle

\IEEEdisplaynontitleabstractindextext

% \fi
\IEEEpeerreviewmaketitle

\IEEEraisesectionheading{\section{Introduction}\label{sec:introduction}}

\IEEEPARstart{M}{any} large datasets are inherently uncertain due to noise, incompleteness, inconsistency, and lack of a sufficient number of training samples which can bring considerable challenges to the analysis process. Using augmentation and imputation methods can improve the quality of data and increase the variety of samples utilised in machine learning models. Novel machine learning techniques such as generative deep learning models are increasingly used in developing data processing pipelines. However, these model require large training sets and often do not generalise well on small-scale datasets, especially non-imaging data with less intrinsic spatial (or temporal) correlations. This hinders the application of generative models in clinical studies and healthcare applications in which imputing missing values and augmentation of the data play a key role in their analysis. Healthcare data may include sensitive information that leads to constrained access to raw data. These issues have a significant impact on the outcomes of the machine learning techniques by misleading or biasing the final results \cite{Surendra2017,Snoke,Charu}. However, Bayesian inference with Hamiltonian Monte Carlo offers a very efficient way to process high-dimensional and small sample datasets. 
In this paper, we apply our proposed Hamiltonian model to a dataset collected from 1342 cancer patients who self-reported their symptom experience during chemotherapy by a team in the School of the Nursing University of California {\cite{PAPACHRISTOU2018}}. We refer to this cancer symptom dataset as the USCF dataset in this paper. The USCF dataset has several missing values and sensitive information. The missing values affect data analysis and provide potentially biased predictions. It is essential to address the missing value problem to improve the predictions while synthetic data generation leads to preserving the data privacy.

Several existing techniques provide data imputation  \cite{RUBIN1976} or synthetic data generations \cite{Snoke}, which resemble essential characteristics of the original data to create new or substitute data. These techniques have their specific strengths and weaknesses, depending on the particular requirements. For instance, \cite{Lin2006} used semantic graph-based synthetic data generation; however, it needs a large number of rules to be defined by the user.  An interesting approach is proposed by \cite{Zhang2017} that uses a differentially private method for releasing high dimensional data using Bayesian networks. Such networks grow extensively as the data dimension increases. Another method proposed by \cite{Lu2014} is a two-phase differentially private data generation mechanism. Nevertheless, the processing time increases dramatically for the higher dimensional and large dataset as the employed perturbation technique is a repetitive method.

Examples of some imputation techniques include record discarding, hot-deck imputation \cite{Andridge2010}, kernel-based imputation \cite{Zhang2011b} and regression-based imputation \cite{Devroye1994, Shichao2005} applicable to different data types to build imputation such as NIIA \cite {Zhang2011b} and DIM \cite{Xiaofeng2019}, and also methods such as support vector machines \cite{Pelckmans2005}. However, they usually suffer from bias caused by dropping cases or replacing data with seemingly suitable values. Hence, the existing approaches in missing data imputation are not ideal for high dimensional healthcare data with a smaller number of samples.

Moreover, most of the approaches perform inadequately for multivariate, high dynamic data. Hamiltonian Monte Carlo (HMC) provides a more practical approach to process the cross-dimensional relations. It uses random walk and Hamiltonian dynamics to adapt to posterior distribution and generates samples in large-scale data streams while exploring the target density and scaling more competent to higher dimensions \cite{Wang2013}. Using HMC, a model can generate multiple samples that mimic data distribution suitable for training and generalising the learning process. In particular, this approach is very beneficial in healthcare scenarios with fewer samples in which the reproductions of the experiments are either very difficult or not feasible. In this paper, we propose an HMC technique capable of updating the momentum value during the inference, which results in better sample generation. 

The contributions of this paper' include proposing a F-HMC sampler with Bayesian inferences to adapt to the posterior of data, generating new samples from the posterior to enrich the quality of data by imputing missing values. We also demonstrate how the proposed work can offer an alternative approach to preserve privacy by augmenting the original data for research analysis.

The remainder of this paper is organised as follows: Section 2 describes the related work. Section 3 details the proposed method and explains the mathematical details of the algorithm. Section 4  presents the experiments and performance evaluations of using the proposed method on the USCF cancer symptom management dataset and MNIST dataset compared to common models in terms of classification performance and augmentation performance called propensity metric. Section 5 concludes the paper and discusses future work.

%\IEEEPARstart{T}{his} demo file is intended to serve as a ``starter file''
%for IEEE Computer Society journal papers produced under \LaTeX\ using
%IEEEtran.cls version 1.8b and later.
%I wish you the best of success.

 

\section{Related Work}\label{sec:RelatedWork}
Handling of missing values and augmentation of experimental data have attracted great attention in healthcare applications. One way of handling the incomplete data is merely by omitting the missing portion from further analysis \cite{Manly2015}. However, deleting rows of data leads to a loss of valuable information. Alternatively, missing values can be predicted based on a close estimation of the data \cite{Li2015, Newgard2015}. e.g., mean value imputation (MVI) \cite{Li2015}, last observation and K-nearest neighbour (KNN). These techniques are considered as suboptimal due to the possibility of their biased results \cite{Newgard2015}. Additional techniques with more accurate predictions have been developed,  including Maximum Likelihood (ML) based methods, Hot Deck Imputation and Multiple Imputation (MI) \cite{Newgard2015}. For instance, MissForest is an iterative imputation method based on Random Forest technique which averages over many classification and regression trees to establish a multiple imputation \cite{Stekhoven2011}. Multiple Imputation using Chained Equations (MICE) \cite{Azur2011}, is another example of a MI technique that is applied widely in healthcare domain for handling missing data. The Expectation-Maximization (EM) algorithm to estimate values of missing data points using Probabilistic Principal Component Analysis (PPCA) estimates missing values by recovering data from its dimension-reduced form \cite{Tipping1999}.
An interesting approach for handling missing data is the Nested Gaussian \cite{Imani2018} process for high-dimensional tensor data which effectively handles the imputation of incomplete tensor data under uncertainty while improving the effectiveness of biomarker extraction, patient monitoring, and decision making. 
Moreover, there is an immense need for enabling data sensitive and private data holders to share their data with scientists by preserving privacy while protecting certain statistical information or relationships between attributes in the original data. However, it is not uncommon to encounter data uncertainty and incompleteness, which adversely affects decision-making in machine learning and more importantly, in healthcare decisions. Data augmentation which coincides essential components from the original dataset but creates new, alternate data is a potential solution to protect privacy. One of the most appealing methods for data augmentation and generating dataset is generative adversarial networks (GANs) \cite{Goodfellow2014}, a type of neural network that produces new data from scratch feeding random noise as input. It can provide realistic (synthetic) copies of the real data as the training progresses.  Unfortunately, GANs are hard to train, likely to have a slow process and unwarranted convergence. GANs also suffer from the problem of vanishing gradient and stocking in the collapse mode \cite{Choi2017,Joshi2019}.
Imputation and augmentation steps are traditionally convoyed individually. The enrichment of data needs in healthcare studies often needs both; therefore, in this work,  we focus on producing substitute data that closely resembles real data in data-limited situations and imputes the missing values using an HMC generator. 
Sampler methods such as Gibbs and MCMC to estimate parameters of interest under missing value are the closest techniques to our proposed model \cite{Wei2018, Yubin2013}. They are based on setting a fixed parameter, the Glmnet\footnote{See
\url{http//web.stanford.edu/~hastie/glmnet/glmnet_alpha.html}} and require more computation time. Our proposed offers a novel sampler to estimate missing values directly instead of estimating the model parameters. The method will also enrich the sampler to explore high dimensional data for estimating missing values while the generated samples preserve data privacy. The details of the model are presented in the method section.

\section{Method for Imputation and Augmentation}
In the literature, sampling methods for data imputation or augmentation estimate parameters of interest under challenging conditions such as missing data or when underlying distributions do not fit the assumptions of Maximum Likelihood processes \cite{MCMC2,MCMC3}. However, this work aims to find a posterior distribution in Bayesian analysis that can be used to estimate the missing data directly. The missing values are modelled by marginalising over generated data in the folded samplers. The advantage of using a Bayesian imputation is the reduction of the credibility interval's width and propagating the uncertainty in the missing values. 
\subsection{Bayesian Approach}
The Bayesian approach provides a framework for making inferences with incomplete data by considering the full-data model as the posterior.
Let $Y= (Y_{ij})$ denote a rectangular data set where $i$ is the data sample, and $j$ is variables' features. Let's partition Y into observed and missing values, $Y= (Y^{obs},Y^{mis})$. Let $M$ be as the mask vector which indicates observed components in $Y_{ij}$ is defined as:
\begin{equation}
   M_{ij}=  \begin{cases} \label{Mask}
             1, & \text{$Y_{ij} $ $ missing $ }.\\
              0, & \text{$Y_{ij} $  $ observed$}.
                  \end{cases}
\end{equation}
%0 : Y_{ij} observed ; 1  Y{ij} missing$
 One can specify the full-data response by calculating the joint model where $w$ is an unknown parameter which consists of $\theta$ and $\phi$. Then the joint model (likelihood) of the full data is
\begin{equation}
  p(Y,M|\theta,\phi) =p(Y^{obs}, Y^{mis},M|\theta,\phi)
  \label{jointModel}
\end{equation}
The joint model in Eq. \ref{jointModel} cannot be evaluated in the usual way because it depends on missing data. However, the marginal distribution of the observed data can be obtained by integrating out the missing data.
Consequently, the joint model  can be written as follows after applying the conditional independence assumption and selection model factorisation:
\begin{equation}
\begin{split} 
  &p(Y^{obs},M| \theta, \phi) = \\
  &\int p(Y^{obs},Y^{mis},M| \theta, \phi)dy^{mis} = \\
  &\int p(M|Y^{obs},Y^{mis}, \theta)p( Y^{obs}, Y^{mis} | \phi)dy^{mis}  \label{bayesian equation}  
\end{split}
\end{equation}

By estimating the integral in Eq. \ref{bayesian equation}, one can determine full data response and consequently generate new samples to utilise for data imputation and augmentation. We show that Monte Carlo samplers, especially F-HMC samplers, are an effective method for the Eq. \ref{bayesian equation} estimation in high-dimensional data.
\subsection{F-HMC Algorithm for Imputation and Augmentation}
We consider the problem formulation as a \textit{d}-dimensional space that X = ($X_1,...,X_d$) is a random variable selected from it.
We consider $M = (M_1,...,M_d)$ as a mask vector which identifies the missing values in the dataset D as defined in the equation \ref{Mask}.
We also define dataset $D = \{(x^i, m^i)\}$, where $m^i$  is the obtained realisation of $M$ corresponding to $x_i$.
Our goal is to impute the unobserved values in each $x^i$.

Given a dataset $D$ and a mask vector $M$, each missing data considered as an unknown parameter which their possible values are drawn and directed to $\bigtriangledown(-\log(P(X|D))$, i.e. the potential energy in Hamiltonian Monte Carlo semantic.
HMC first models the posterior distribution of each feature dimension of data using the Gaussian likelihoods with a Laplacian prior, to find the $mu$ and $sigma$s of feature distribution. Then, all the $mu$ and $sigma$s for each feature dimension are given to another HMC (fold) with respect to the cross-correlation between all features. The F-HMC adopts the results using gradient information to draw samples from the cross-dimensional distribution of features. After a burn-in time, the algorithm converges, and it can generate samples that belong to the posterior of the complete dataset. For imputation, the missing values are replaced by marginalisation over generated samples correspond to that missing part.
Algorithm \ref{alg:HMC} presents the steps in our approach, where $D$ is the incomplete dataset, $M$ is the mask vector to indicate missing values in $D$, $\eta$ is the step size for HMC dynamics, and $k$ is the number of generated samples of the full dataset.

\begin{algorithmic}
%\caption{Proposed data imputation and augmentation algorithm}
\label{alg:HMC}
\STATE \textbf{Input} : {$D, M, k, \eta$}
\STATE \textbf{Output} : {$X^1, X^2, X^3,$... $, X^K$}

\STATE \textbf{Initialisation} : $X^0$
\FOR{$i\leftarrow 1$ to $k$}
\STATE $X_0 \gets X^{k-1}$
\STATE $mu_d,$ $sigma_d \gets HMC$ for each
\STATE $x_1,x_2,\dots,x_d$ separately;
\STATE $MU,$ $SIG \gets$ joint $mu_d,sigma_d$
\STATE $ X_{k-1} \gets$ $HMC(mu_d,sigma_d,\eta)$
\STATE $X_k \gets X_{k-1} \oplus M$
\STATE $X^k \gets X_k$\;
\ENDFOR
%\For{$i\leftarrow 1$ \KwTo $k$}{
%    $X_0 \gets X^{k-1}$;
    
%    $mu_d,$ $sigma_d \gets HMC$ for each $x_1,x_2,\dots,x_d$ separately;
    
%   $MU,$ $SIG \gets$ joint $mu_d,sigma_d$
    
%   $ X_{k-1} \gets$ $HMC(mu_d,sigma_d,\eta)$
    
%   $X_k \gets X_{k-1} \oplus M$
    
%  $X^k \gets X_k$\;
%}
\end{algorithmic}

After initialisation of $X_0$ with white noise, the algorithm starts. First HMC iterates in parallel over each feature dimension of $D$ to calculate $mu_d, sigma_d$, separately. Next, it produces new $MU$ and $SIG$ by concatenating the $mu_d, sigma_d$ which allows the cross-correlation of features to be considered in the learning. Then $MU$ and $SIG$ are given to the second HMC sampler to explore the posterior and return samples from that posterior. Operator $\oplus$ is for marginalising generated samples over missing values with the help of matrix $M$.
The algorithm keeps a record of the latest state of outputs for the next round to make it more accurate over time. The outputs of the algorithm $X^1,X^2,\dots,X^k$ are the complete datasets drawn from the estimated posterior. The algorithm can impute missing values and generate more samples from posterior, in case of requiring more samples from the data.

\section{Experiments and Evaluations}
\subsection{Experiments Setup}
We implemented the algorithm \ref{alg:HMC} using Pymc3 \footnote {https://docs.pymc.io/} and the baseline methods using the available R packages. We ran all the experiments ten times and reported performance metrics such as accuracy, precision and recall along with standard deviation across ten experiments for the evaluation purpose.
The experiments ran on two datasets: (1) the MNIST dataset was considered as a high dimensional common dataset to compare several baseline methods with the proposed approach to illustrate that the proposed method is generalisable to other datasets and (2) the USCF dataset; Cancer symptom management dataset collected from cancer patients during chemotherapy. 

To evaluate the proposed method's performance in data imputation and augmentation, we have intended three measurement levels: distance metric, outcome performances on classification and propensity metrics. The distance metric is employed in scenarios that we are aware of the value of missing parts. The missing parts are generated randomly and dropped from the dataset. Then given dataset with artificial missing values, the proposed model is used to impute them, and distance metric Normalised Root Mean Square Error (NRMSE) is used to show the performance of imputation. 

In the UCSF dataset that we are not aware of the value of missing parts, the evaluation is done based on the classification performance of data after applying imputation and augmentation techniques. The more improvement over classification stands for higher enrichment in the quality of data hence more reliable imputation technique.

Finally, the propensity metric is utilised to focus only on augmentation to evaluate the goodness of the synthetic data in resembling the original data. 
The implemented algorithm codes are available online at
\url{https://github.com/nargesiPSH/Folded-Hamiltonian-Monte-Carlo} 

\subsection{MNIST Data}
\textsf{MNIST}\footnote{\url{ http://yann.lecun.com/exdb/mnist/}} dataset containing 60,000  images of handwritten digits of 28*28 pixels is chosen to evaluate data imputation and augmentation performance on a standard dataset. We consider each pixel as a feature dimension; therefore, the data are converted to the size of 60000 by 784 dimensions. To generate artificial missing values in MNIST, we have dropped pixels from images randomly, and then, we reconstructed the images using the proposed approach and several baseline methods for comparison. 

\begin{figure}[h]
\centering
\includegraphics[width =0.9\linewidth]{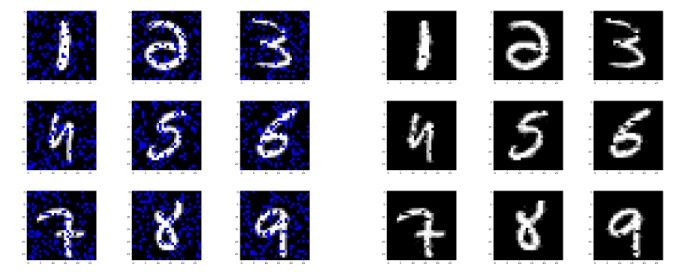}
\caption{Image reconstruction on MNIST dataset using the F-HMC for data imputation after dropping $20\%$ pixels as missing values (blue pixels).}
\label{MNISTReconstruction}
\end{figure}

Figure \ref{MNISTReconstruction} shows reconstructed images from MNIST dataset after inserting $20\%$ missing data; on the left side the blue pixel resembles random dropped pixels, and on the right side reconstructed image using the proposed model is shown.

\begin{figure}[h]
\centering
\includegraphics[width =1.0\linewidth]{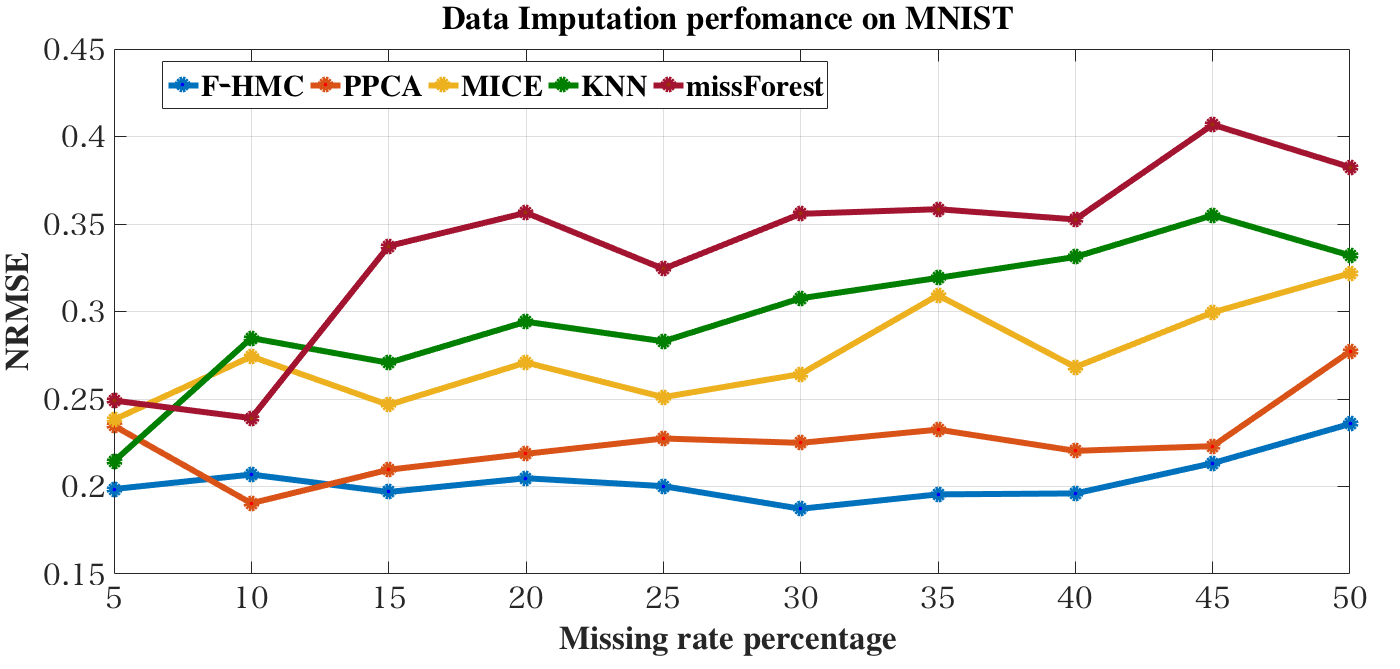}
\caption{Comparison of various data imputation techniques; KNN, MICE, missForest, and the F-HMC on MNIST dataset}
\label{MNISTComparison}
\end{figure}

We compared the imputation performance of the proposed approach with baseline methods, namely, KNN, missForest, PPCA, and MICE in term of NRMSE as distance metric defined as follow:
\begin{equation}
    \sqrt{mean((X_{true}-X_{imp})^2)/var(X_{true})} \label{eq:nrmse}
\end{equation}

where $X_{true}$ is the true data matrix, $X_{imp}$ is the imputed data matrix, and $’mean’/’var’$ is used as a short notation for the empirical mean and variance. As Shown in Figure \ref{MNISTComparison}, our approach outperforms baseline methods considering various levels of missing rate in the data.
\begin{figure}[h]
\centering
\includegraphics[width =1.0\linewidth]{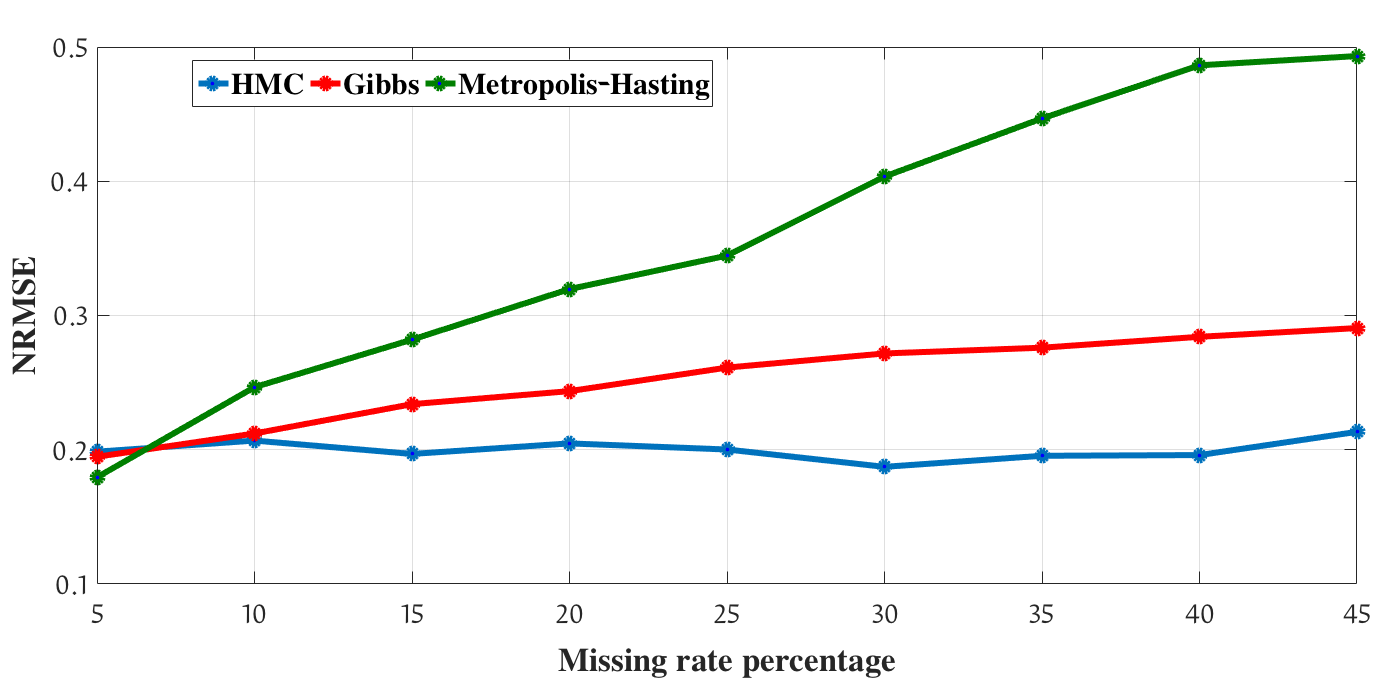}
\caption{Comparing the impact of sampler's type on data imputation performance using the proposed approach on MNIST dataset}
\label{MNISTCompareMCMC}
\end{figure}

We also explored the impact of using various Monte Carlo sampler such as classic Metropolis-Hasting and Gibbs sampler on the outcome of the method in Figure \ref{MNISTCompareMCMC}, which shows that the Hamiltonian sampler is exceeding the others in terms of NRMSE.
\subsection{The USCF Dataset}
We have evaluated the proposed approach on a clinical study dataset that focused on analysing the symptoms reported by cancer patients during chemotherapy. The dataset was collected by a team in the School of Nursing at the University of California, San Francisco (UCSF) \cite{PAPACHRISTOU2018}. In this paper, we have used an anonymised dataset from the UCSF study. The original study was approved by the Committee on Human Research at the University of California at San Francisco and by the Institutional Review Board at each of the study sites (No. 10-02882) and written informed consent was obtained from all patients. The symptom instrument is a modified version of the Memorial Symptom Assessment Scale (MSAS) that was used to evaluate the occurrence, severity, frequency, and distress of 38 symptoms commonly associated with cancer and its treatment {\cite{Papachristou20182}}. Table 1 shows the 38 symptoms and the rate of missing values in the data. Patient in this study has four types of cancers (i.e. breast cancer, gastrointestinal cancer, gynaecological cancer and lung cancer) that represented 40.2\% , 30.1\%, 
,17.6\%,  and 12.1\% of data, respectfully and consists of 1,044 female and 298 male patients. 
\begin{figure}[h]
\centering
\includegraphics[width =0.9\linewidth]{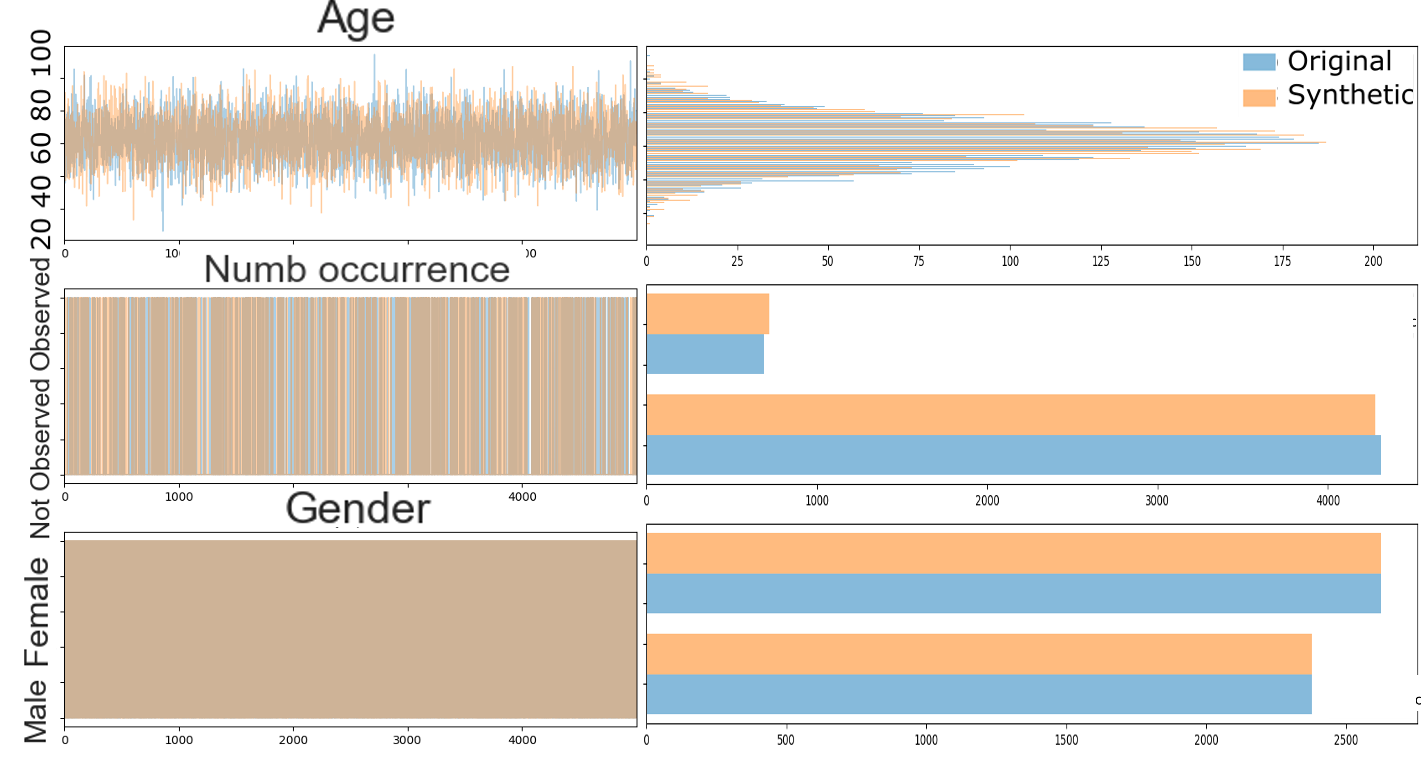}
\caption{Three selected symptoms form the USCF dataset during the 5000 iterations of the model. The left-side shows the distribution of each generated symptoms, and the right-side chart shows the path that samplers went through to generate the corresponding symptoms.}
\label{USCiSimulation}
\end{figure}
After running our F-HMC approach on the USCF data, Figure \ref{USCiSimulation} is generated to exhibit the trace of the model. It shows three symptoms, namely age, cough occurrence and gender selected form USCF dataset during the 5000 iterations of the model. The left-side shows the distribution of each generated symptoms, and the right-side chart shows the path that samplers operated to generate the symptoms. It shows that the model accommodates different data types and the spreads the expected range of real original data due to the selection process of HMC.

The experiments on USCF data consists of three parts: (1) measuring the imputation performance of our proposed approach in comparison with baselines and (2) validating the quality of data augmentation in the proposed method and (3) propensity measurement.
\subsubsection{Data Imputation}
Missing values in this study were imputed by the F-HMC approach and compared with the baseline methods.  Unlike the MNIST experiments, we do not have access to the actual value for missing parts in the UCSF data as they happened naturally throughout the clinical data collection; therefore, NRMSE can not be applied as a metric to report data imputation performance. 
We evaluated the performance based on the classification metric after applying imputation techniques, including baselines and the proposed method on each cancer type. We have reported the results after 10-fold cross-validation by dividing the dataset into train and test set before running the sampler on them. Since it is a multi-class classification problem, the value for each metric is reported in macro-averaging, which averages the performances of each individual class. Higher classification metric corresponds to more solid imputation performance. It should be noted that measuring the classification performance on estimating cancer type based on symptoms is not the goal of this study. Still, it provides a way to measure the impact of imputation methods on data analysis.

\begin{figure}[t!]
\centering
\includegraphics[width =1.0\linewidth]{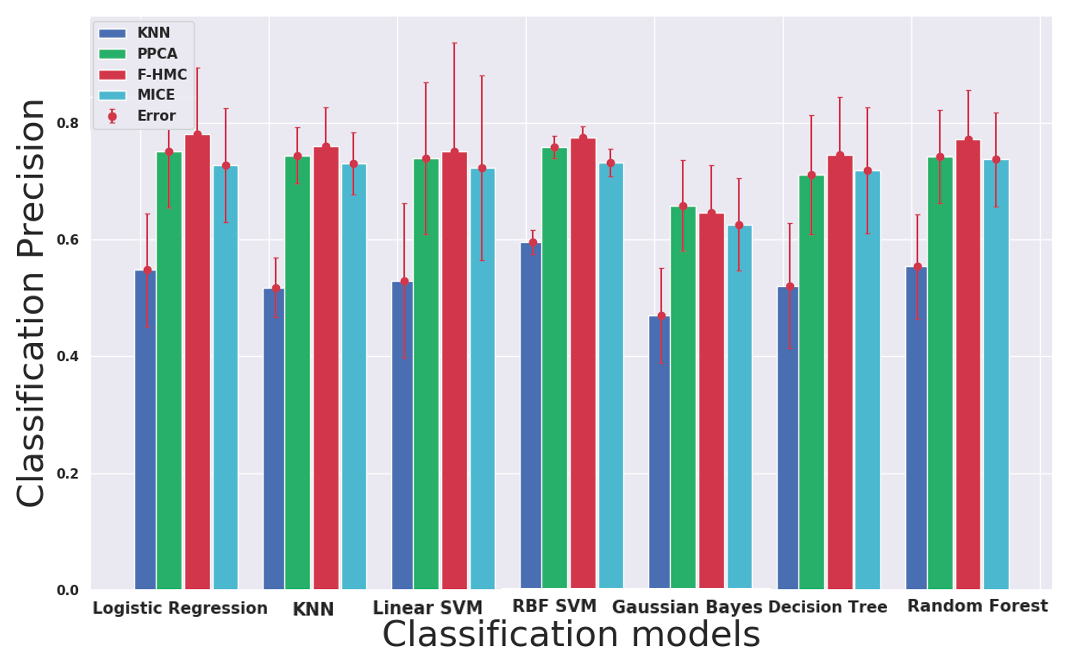}
\caption{Classification performance of various machine learning techniques on an imputed version of the USCF data after applying KNN, PPCA, F-HMC, MICE. A higher score indicates a more reliable imputation method.}
\label{USCimputationCompare}
\end{figure}

Figure \ref{USCimputationCompare} shows that the predicted precision in the data imputed by the proposed approach is higher than the baseline methods and confirms that imputation using the suggested approach is more substantial than baselines.
\subsubsection{Propensity Measurement}
\begin{figure}[t!]
\centering
\includegraphics[width =1.0\linewidth]{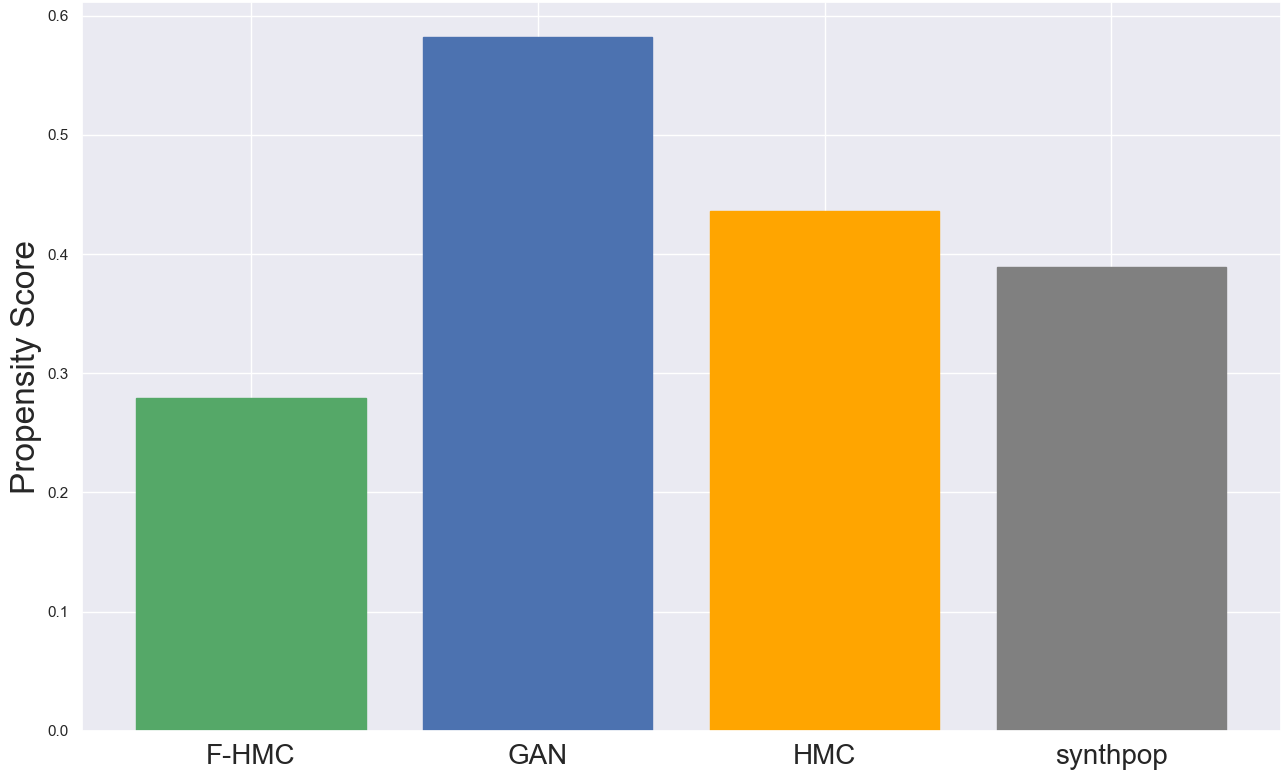}
\caption{Comparing propensity score in data augmentation on the USCF dataset as a metric to evaluate the quality of synthesised data. A lower score indicates higher quality.}
\label{propensity}
\end{figure}

\begin{figure*}[t!]
\centering
\includegraphics[width =1.0\linewidth]{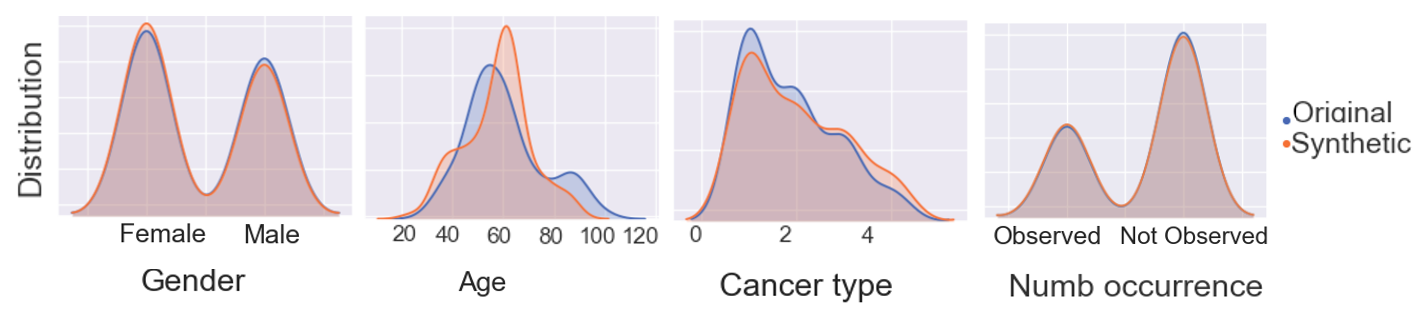}
\caption{Probability distribution of four randomly selected symptoms from the USCF dataset. The blue line shows the original data, and the red line shows the augmented data using the proposed model.}
\label{pairplot}
\end{figure*}
\subsubsection{Data Augmentation for Sensitive Data Sharing}

\begin{figure*}%
    %\centering
    \subfloat[\centering Recall]{{\includegraphics[width=0.5\linewidth]{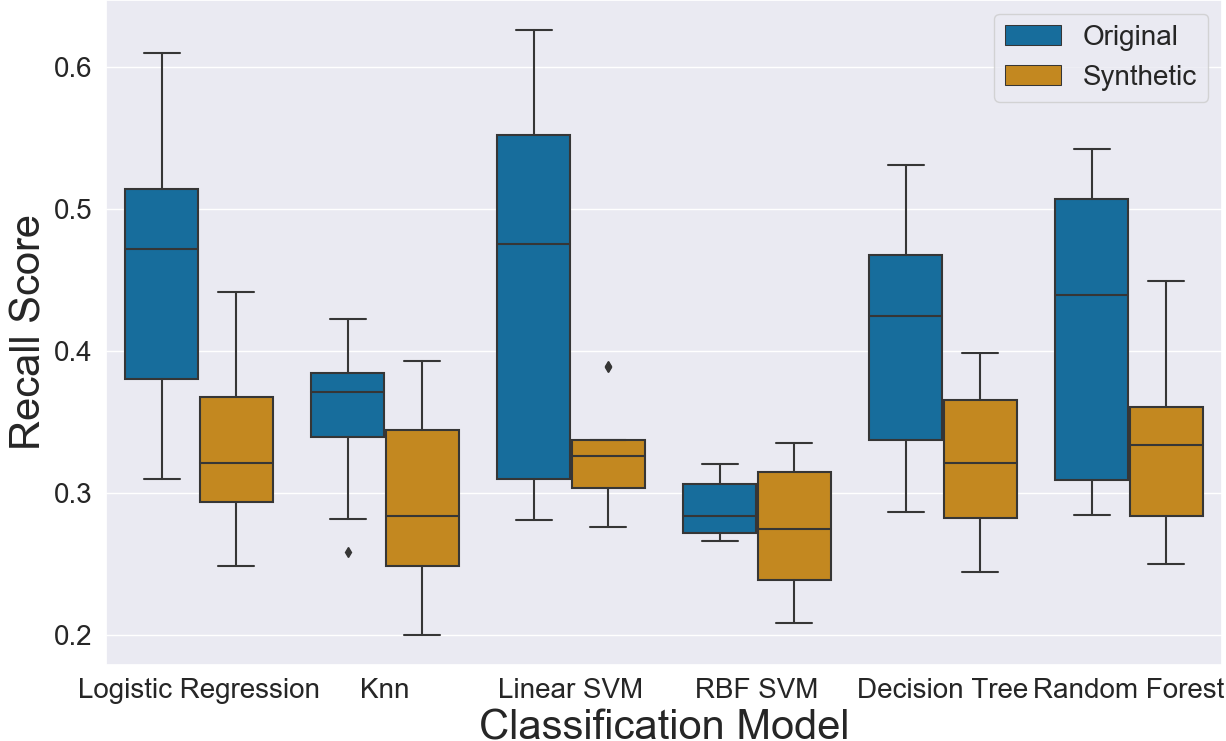} }}%
    \qquad
    \subfloat[\centering Accuracy]{{\includegraphics[width=0.5\linewidth]{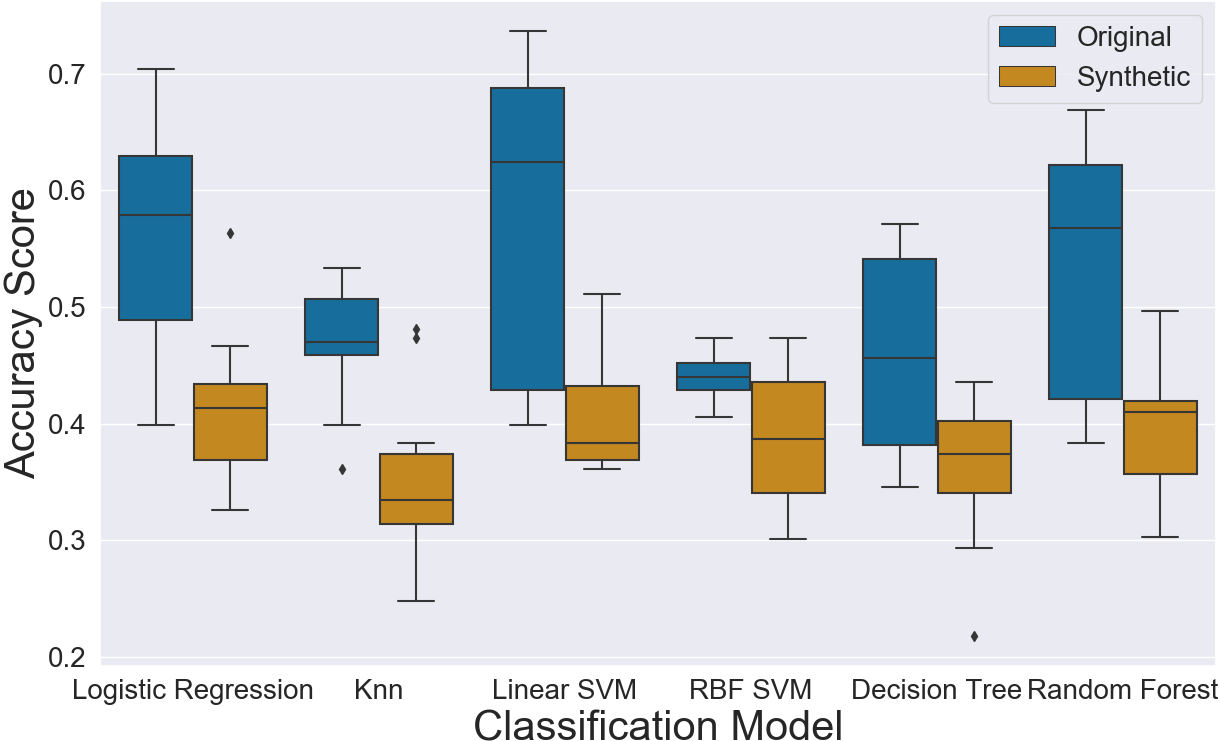} }}%
    \caption{Reproducibility of classification performance on original and synthetic USCF data using the proposed approach after applying various classification models}%
    \label{accuracyrecall}%
\end{figure*}

As we mentioned earlier, the proposed approach is also capable of augmenting data in privacy-preserving use-cases. Figure \ref{pairplot} presents the agreement in the visualisation of selected features in both original and synthetic data using the F-HMC approach. Beside similarity in visualisation, it is expected that certain ML methods perform similarly on both original and augmented data considering uncertainty. We judged the performance based on the accuracy, precision, recall, and F1 score of data classification on each cancer type reported in macro-averaging which averages the performances over each individual class.
As shown in Figure \ref{accuracyrecall}, each method's performance is reproduced on the synthetic data. For instance, Logistic Regression has higher accuracy than nearest neighbour in both original and augmented data Figure \ref{accuracyrecall} section (b). Another example is in the metric recall, where Kernel SVM recall is less than recall in the decision tree on original data that agrees with the same results on synthetic data. The same agreement can be seen in the results of both original and synthetic data in Figure \ref{accuracyrecall}.

\begin{figure*}%
    %\centering
    \subfloat[\centering F1-score]{{\includegraphics[width=0.5\linewidth]{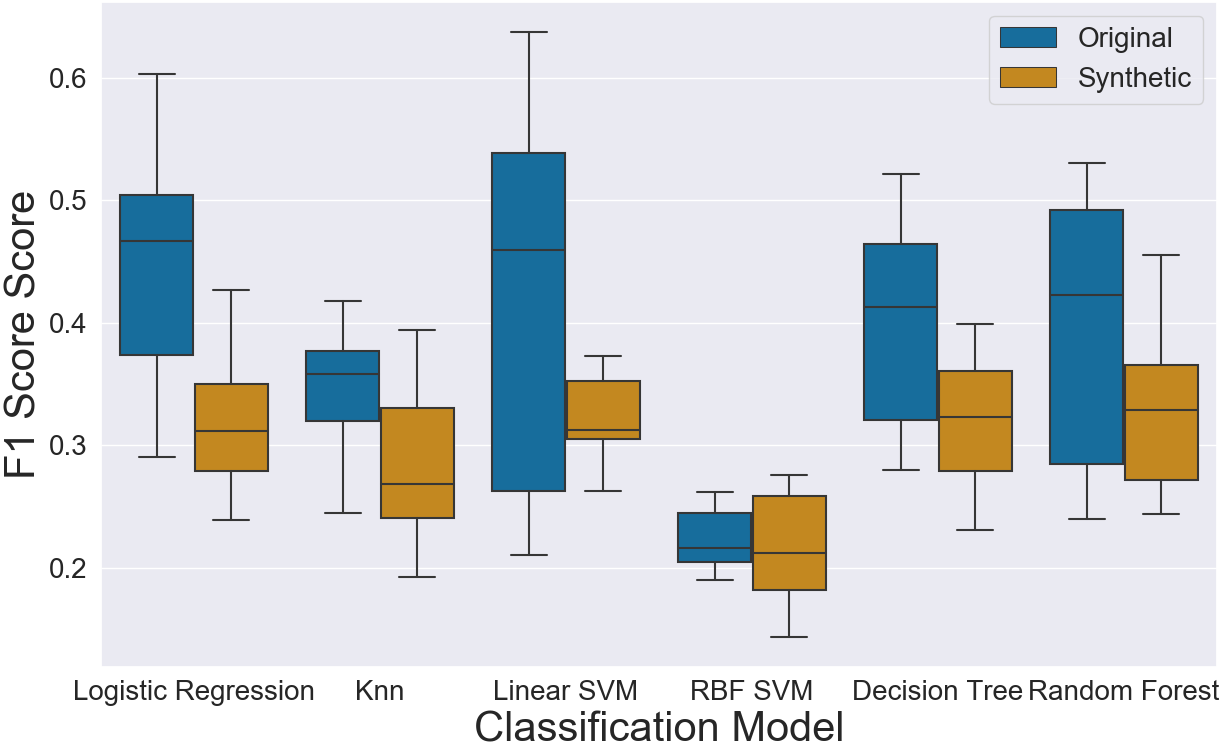} }}%
    \qquad
    \subfloat[\centering Precision]{{\includegraphics[width=0.5\linewidth]{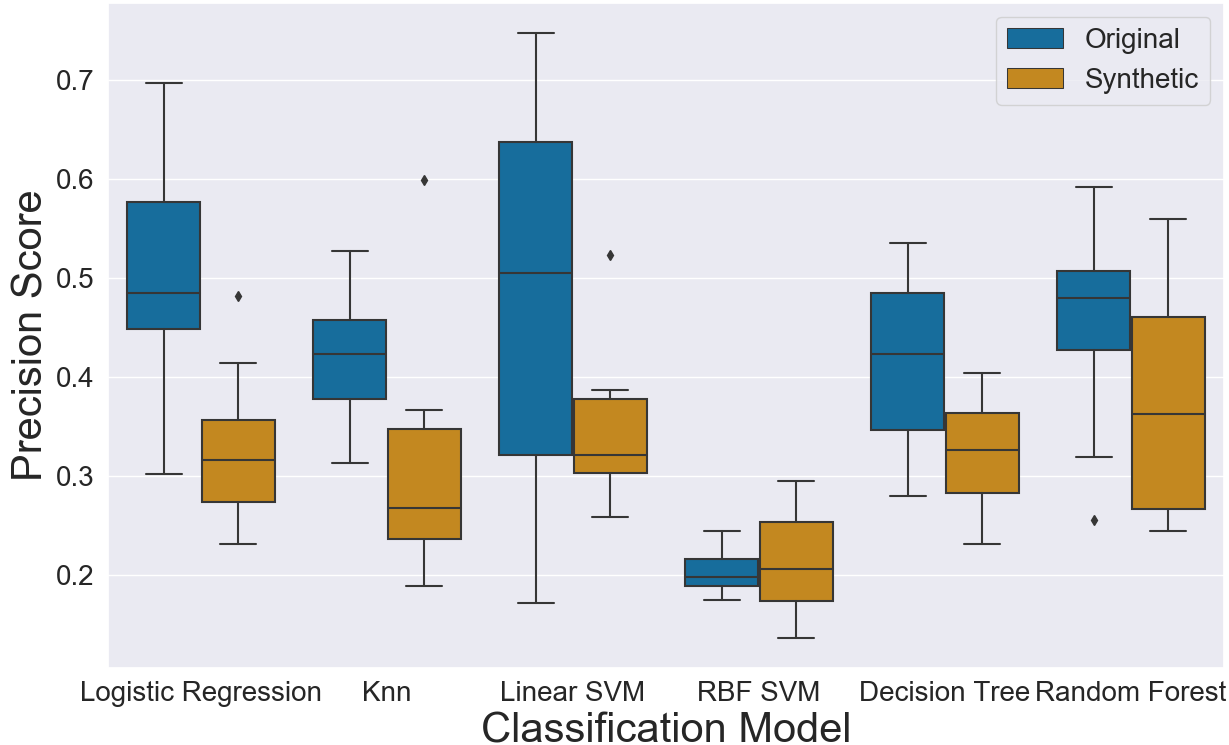} }}%
    \label{f1precision}%
\end{figure*}

\begin{figure*}%
    %\centering
    \subfloat[\centering Original data]{{\includegraphics[width=0.5\linewidth]{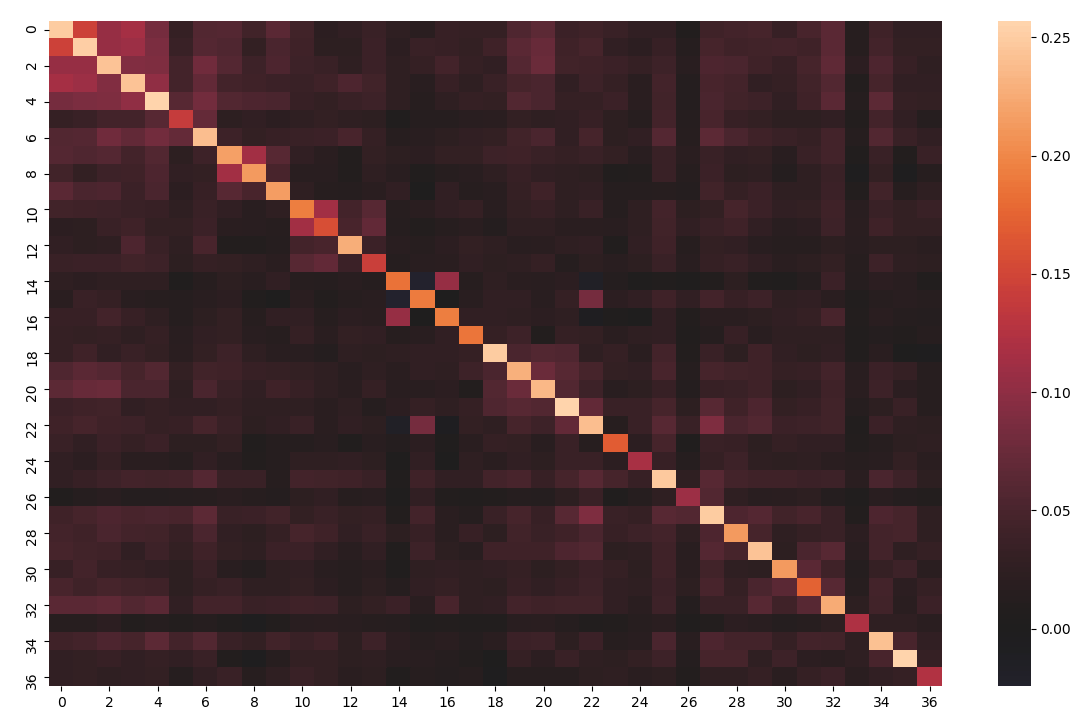} }}%
    \qquad
    \subfloat[\centering Synthetic data]{{\includegraphics[width=0.5\linewidth]{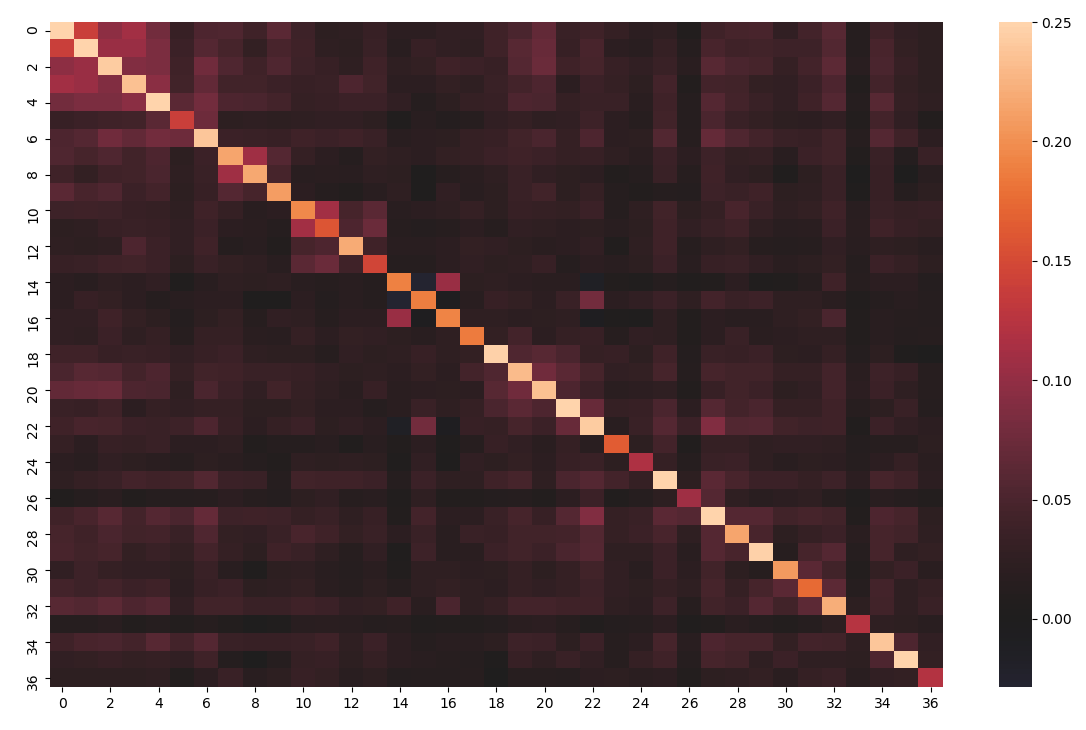} }}%
    \caption{Cross relationship between symptoms}%
    \label{crossrelation}%
\end{figure*}

Another aspect of the work is the cross-relationship between the features in original and synthetic data. We have measured covariance of the symptoms in patients in original and synthetic data. Figure \ref{crossrelation} shows that the cross-relationship between the features of synthetic data is comparable to the original data. The x and y axes show feature numbers listed in table 1 in the Appendix.

We have evaluated the quality of synthetic data by measuring propensity score and comparing it with baseline methods such as synthpop \cite{synthpop}, GAN, F-HMC and basic HMC.
The propensity score is the probability of a given data point being assigned to a particular class. Therefore the higher quality of synthetic data leads to more problematic for a classifier to distinguish between real and synthetic datasets \cite{Snoke2016}. As Figure \ref{propensity} shows, the synthetic data generated by the F-HMC approach has the lowest propensity score meaning the higher quality of data. In Figure \ref{propensity}, GAN is not performing adequately due to lack of sufficient number of samples required for training and converging the network.
\section{Challenges and Discussion}
The sampling and generative model study are often needed to compare a newly invented model and demonstrate that it is better at capturing some distribution than a pre-existing model, which could be an intricate and subtle task. For instance, it is critical to determine clearly what exactly is being measured considering, in most cases, the log-probability of data can be approximated rather than evaluated. However, if the model's performance in practice is preferred, then it is adequate to compare models based on a criterion specific to the practical task of interest. For example, based on ranking criteria such as precision and recall in classification \cite{Goodfellow-et-al-2016}.

Another subtlety of evaluating generative models is that practitioners often evaluate sampling and generative models by visually inspecting the samples. A generative model trained on data with tens of thousands of modes may ignore a small number of modes. The human observer would not easily inspect or remember enough samples and dimensions to detect the missing variation \cite{Goodfellow-et-al-2016}. 
Theis \textit{et al.} \cite{theis2016note} and Goodfellow \cite{Goodfellow-et-al-2016} reviewed many of the issues involved in evaluating generative and sampling models, including many of the ideas discussed above. They highlight the fact that there are many different uses of generative models and that the choice of metric must match the model's intended use. 

In this paper, we faced similar issues in measuring and evaluating the goodness of generated data. Here we discuss the limitation and approach to alleviate them. In any generative model, we need to generate large enough samples to ensure that the model is visited all space sample which is available at our Gitbub \footnote{\url{https://github.com/nargesiPSH/Folded-Hamiltonian-Monte-Carlo/blob/main/USCF_dataset/Data_Augmentation/generated.xlsx}} . The original dataset that we work is already small, and the size of generated samples from the model is significantly larger than that. We start to select from the generated dataset and use them to compare with the original dataset. We noticed that selection itself causes bias. We need to run another selection method to choose from our generated model, which can cause in a never-ending loop. To make it relatively comparable, we decided to add selection bias to both original and generated groups. We have chosen 500 samples in USCF data and 500 samples in F-HMC, and compared results based on that in each experiment.
\label{sec:propose}
\section{Conclusion}
In this work, we propose a F-HMC technique to impute missing values and generate augmented samples in high dimensional but small healthcare datasets. We demonstrated that a hybrid approach of Bayesian inference and F-HMC provides an effective way to augment data samples and impute missing values by including the cross-correlation between the feature dimensions. The work assesses the performance of the proposed method in three stages: i) distance metric (NRMSE); ii) classification score impact (accuracy, recall, precision and f1score); and iii) propensity score on the cancer symptom management dataset (i.e., USCF dataset) and MNIST dataset. We have demonstrated that imputing missing values using our approach outperforms imputation using baseline methods, i.e., KNN, missForest, PPCA, and MICE in terms of NRMSE and performance measures such as accuracy and precision in classification. The quality of augmented data using the proposed method is more reliable than GAN and synthpop in terms of the propensity score. Besides, the outcome on MNIST dataset shows that the model can generalise to other applications. Overall, the proposed technique provides a unique opportunity to improve the quality of data processing for healthcare applications in which data collection is time and effort consuming, and the data is prone to having missing values and quality issues. The proposed method is especially effective when the size of the dataset compared to the number of the given samples is relatively small. The proposed models offer a very effective solution for use-cases in which the popular generative and adversarial neural network models struggle to converge or do not generalise well to multivariate data with small number of training samples. One of the challenges in the work that opens an interesting topic for this research's future is fluctuations in augmented data. The HMC samplers force the proposed model to generate data concentrated on the distribution of original data which cause satisfying augmentation. Still, the quality of augmented data can improve by adding more variations to the generated data.

\appendices
\section{More detail of performance of HMC}
\begin{figure}[h]
\centering
\includegraphics[width =0.9\linewidth]{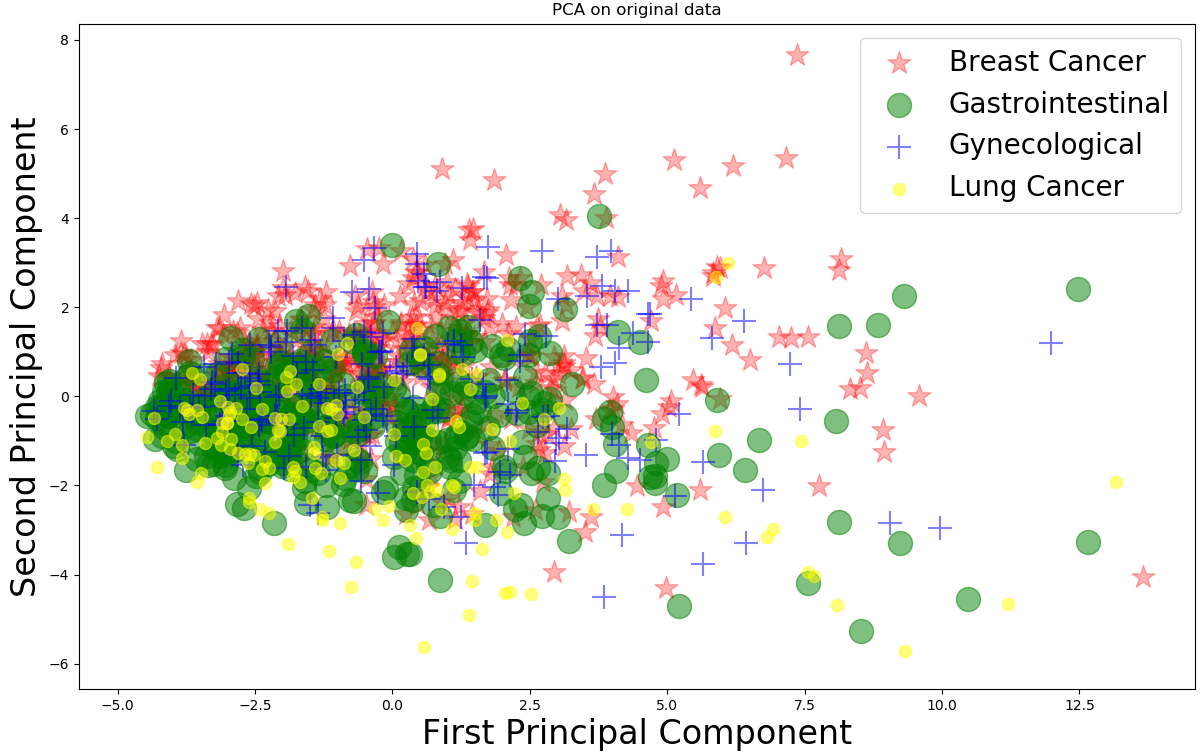}
\caption{PCA comparison of the USCF augmented and original data}
\label{PCA2}
\end{figure}

\begin{figure}[h]
\centering
\includegraphics[width =0.9\linewidth]{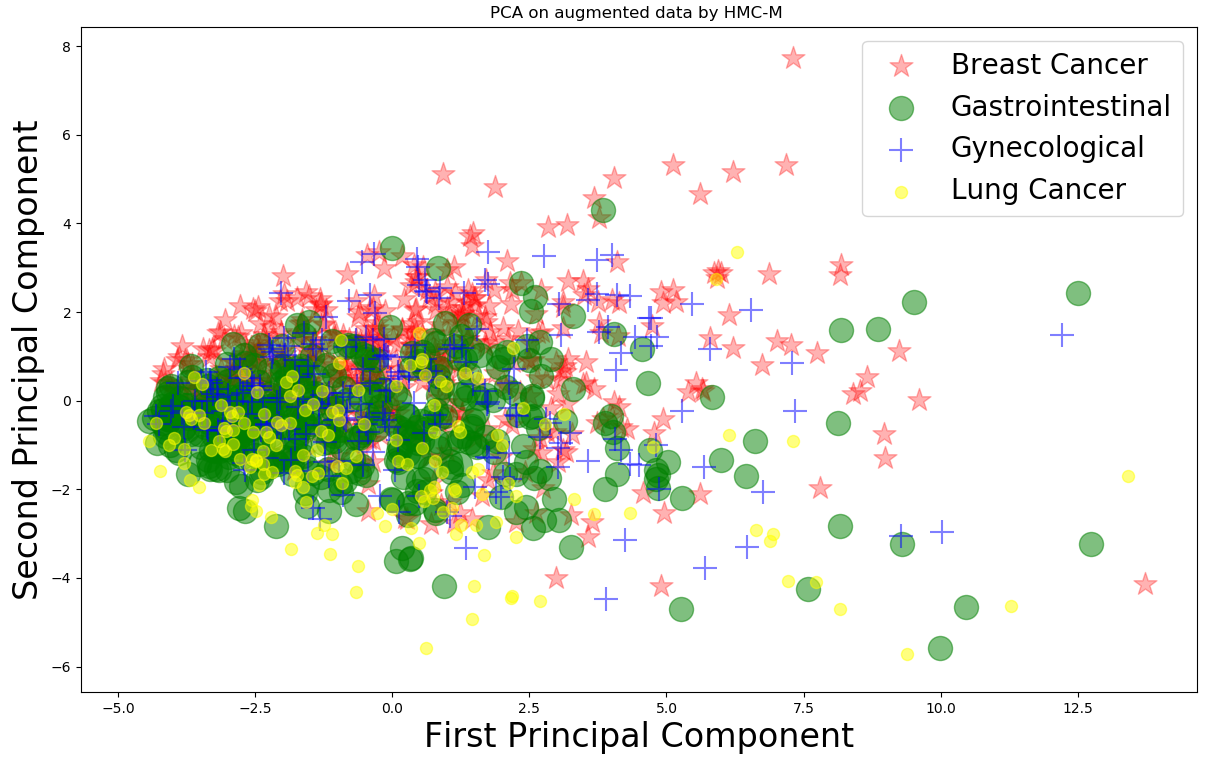}
\caption{PCA comparison of the USCF augmented and original data}
\label{PCA1}
\end{figure}

Table \ref{USCFMissing} shows list of symptoms in the clinical study and missing rate in each symptom. Figure \ref{PCA1} and Figure \ref{PCA2} shows the visualization of the original USCF data and augmented data using F-HMC.

\begin{table*}[]
\caption{The symptoms in the UCSF dataset (n=38) and the ratio of missing values for each symptom}
\resizebox{\linewidth}{!}{%
\begin{tabular}{|l|l|l|l|l|l|}
\hline
Symptom & Missing ratio & Symptom & Missing ratio & Symptom & Missing ratio \\ \hline
difficulty sleeping &  0.0314 &sweats    &0.0275   &weight loss    &0.0236   \\
worrying   & 0.0362  &hot flashes    &0.0275   &increased appetite    &0.0259   \\
feeling sad  &0.0306   &sexual Interest   &0.0283   &itching    &0.0291   \\
feeling irritable   &0.0251   &short breath    &0.0228   &hair loss    &0.0330  \\
feeling nervous  &0.0346   &difficult breathing    &0.0251   &changes in skin    &0.0259 \\
concentrating  &0.0338  &cough    &0.0291   &like myself &0.0251   \\
energy lack  &0.0467   &chest tightness    &0.0220   &food tastes    &0.0236   \\
feeling drowsy   &0.0394   &weight gain    &0.0322  &lack of appetite    &0.0291   \\
mouth sores    &0.0212   &swallowing  &0.0181   &dry mouth    &0.0338   \\
vomiting    &0.0174   &nausea    &0.0370   &constipation    &0.0291   \\
diarrhea   &0.0236  &  cramps &0.0244 & bloated &0.0236 \\
swelling &0.0174  & pain &0.0378 & numbness &0.0402 \\
dizziness    &0.0197   & urination &0.0197 & &\\\hline
\end{tabular}}

\label{USCFMissing}
\end{table*}

\section*{Acknowledgement} Payam Barnaghi's and Samaneh Kouchaki's work is supported by the Medical Research Council (grant no: UKDRI-7002), Alzheimer’s Society and Alzheimer’s Research UK as part of the and Care Research and Technology Centre at the UK Dementia Research Institute (UK DRI).

\vskip 0pt plus -1fil
\begin{IEEEbiography}[{\includegraphics[width=1in,height=1.25in,clip,keepaspectratio]{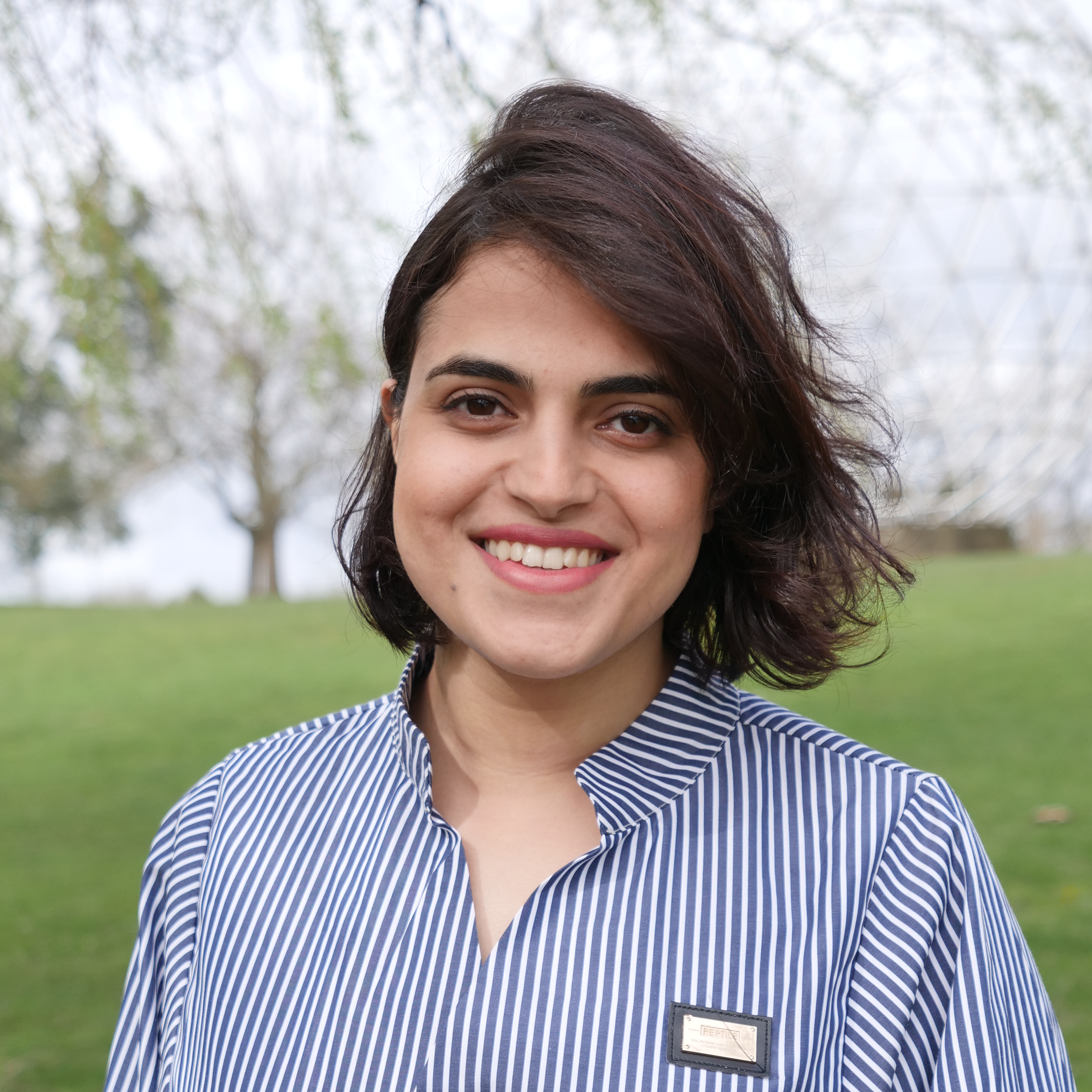}}]{Narges Pourshahrokhi} is currently pursuing her Ph.D.
in Machine Learning at the Centre for Vision,
Speech and Signal Processing (CVSSP) at the
University of Surrey. Her research interests include Internet of Things (IoT), big data analytics, data imputation and augmentation and time-series data processing.
\end{IEEEbiography}
\vskip 0pt plus -1fil
\begin{IEEEbiography}[{\includegraphics[width=1in,height=1.25in,clip,keepaspectratio]{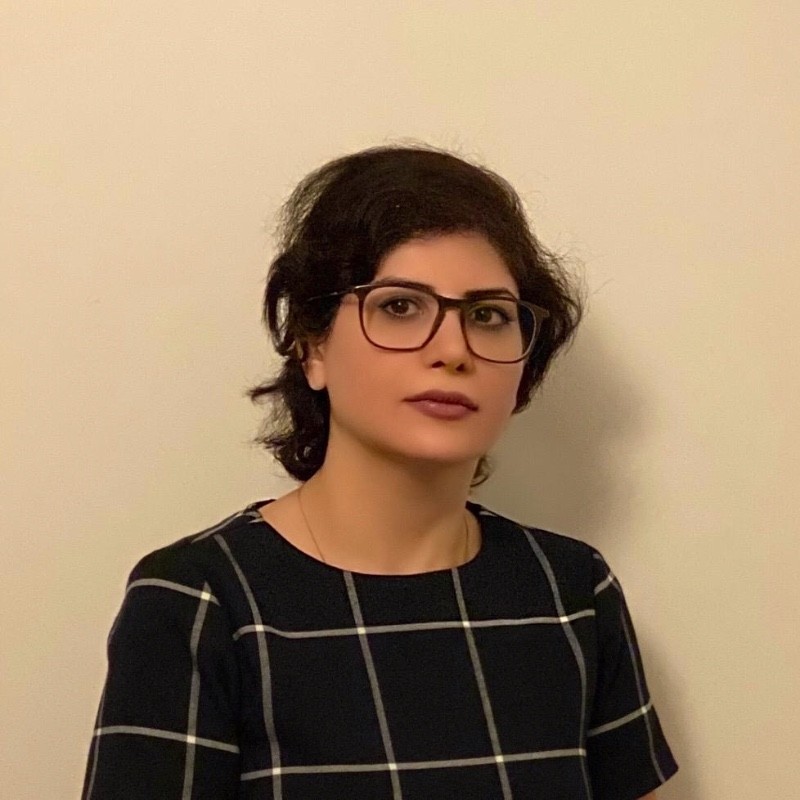}}]{Samaneh Kouchaki} is Lecturer in Machine learning for Healthcare at the Department of Electronic and Electrical Engineering and a member of the Centre for Vision, Speech and Signal Processing (CVSSP) at the University of Surrey. Her research interests include machine learning, health informatics, biomedical signal processing and computational biology. 
\end{IEEEbiography}
\vskip 0pt plus -1fil
\begin{IEEEbiography}[{\includegraphics[width=1in,height=1.25in,clip,keepaspectratio]{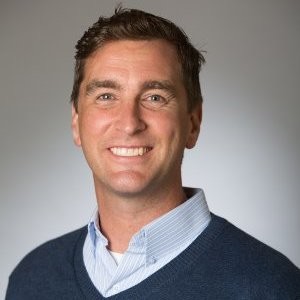}}]{Kord M. Kober}is an Assistant Professor in the Departments o Physiological Nursing at the University of California, San Francisco. His program of research focuses on the investigation of the molecular mechanisms for and prediction of the severity of cancer chemotherapy-related fatigue using a Multi-staged integrated omics approach. 
\end{IEEEbiography}

\vskip 0pt plus -1fil
\begin{IEEEbiography}[{\includegraphics[width=1in,height=1.25in,clip,keepaspectratio]{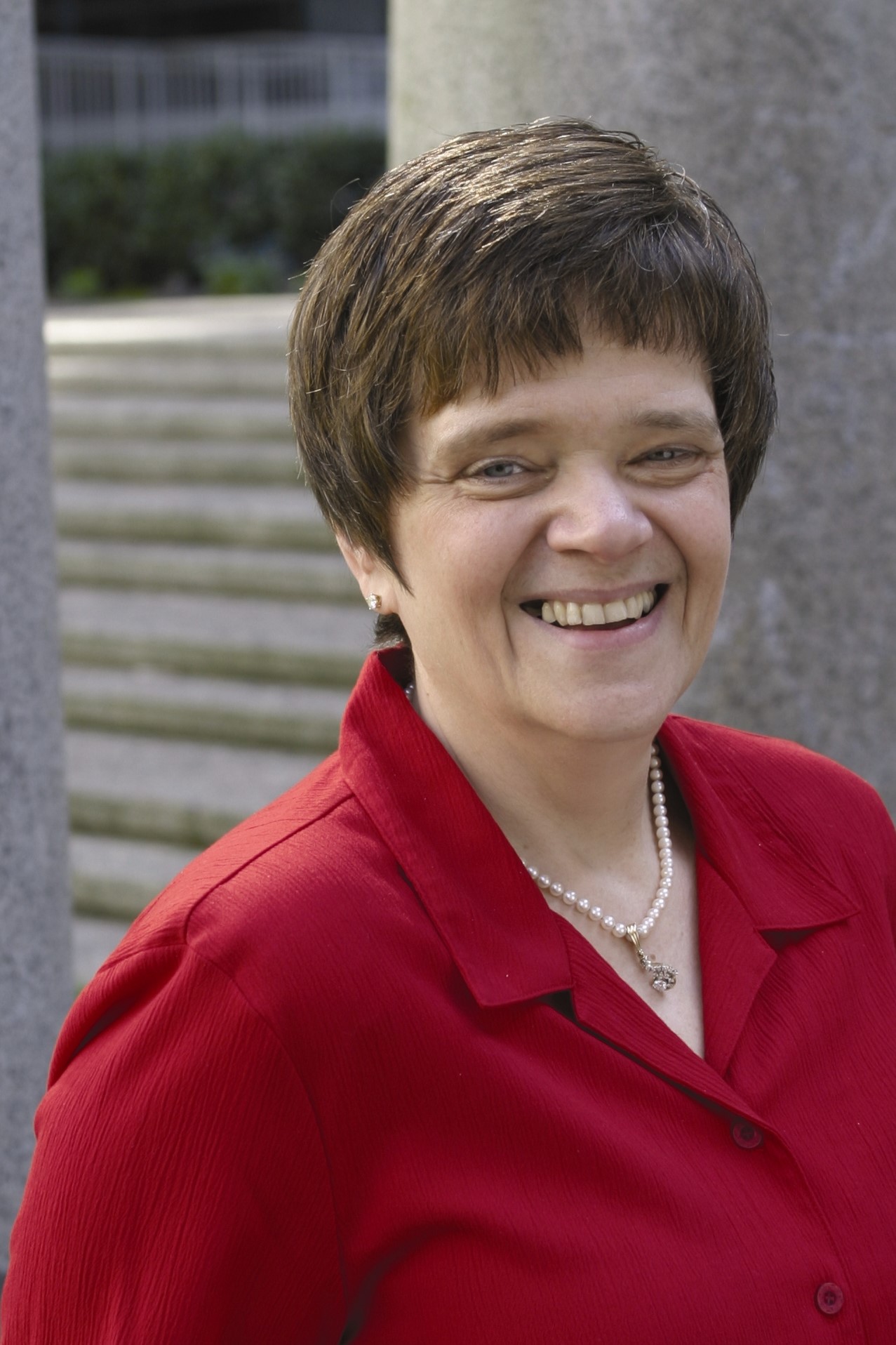}}]{Christine Miaskowski}
RN, PhD is a Professor in the Departments of Physiological Nursing and Anesthesia and Perioperative Care at the University of California, San Francisco. Her program of research focuses on an evaluation of phenotypic and molecular characteristics that place oncology patients at increased risk for a higher symptom burden.
\end{IEEEbiography}

\vskip 0pt plus -1fil
\begin{IEEEbiography}[{\includegraphics[width=1in,height=1.25in,clip,keepaspectratio]{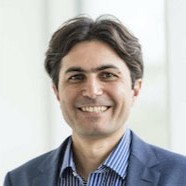}}]{Payam Barnaghi}
is Chair in Machine Intelligence Applied to Medicine in the Department of Brain Sciences at Imperial College London. He is Co-PI and Deputy Director of the Care Research and Technology Centre at the UK Dementia Research Institute (UK DRI). His research focuses on developing machine learning, adaptive algorithms and Internet of Things solutions for healthcare applications. 
\end{IEEEbiography}


%\vfill


\end{document}

\onecolumn
%\title{Bare Demo of IEEEtran.cls for\\ IEEE Computer Society Journals}
\appendices
\section{More detail of performance of HMC}

\begin{figure}[H]%
    %\centering
    \subfloat[\centering PCA comparison of the USCF augmented and original data]{{\includegraphics[width=0.5\linewidth]{images/PCA original .png} }}%
    \qquad
    \subfloat[\centering PCA comparison of the USCF augmented and original data]{{\includegraphics[width=0.5\linewidth]{images/PCA augmented HMCM.png}}}%
    \caption{Cross relationship between symptoms}%
    \label{PCA2}%
\end{figure}


Table \ref{USCFMissing} shows list of symptoms in the clinical study and missing rate in each symptom. Figure \ref{PCA2}  shows the visualization of the original USCF data and augmented data using F-HMC.

\begin{table}[H]
\caption{The symptoms in the UCSF dataset (n=38) and the ratio of missing values for each symptom}
\resizebox{\linewidth}{!}{%
\begin{tabular}{|l|l|l|l|l|l|}
\hline
Symptom & Missing ratio & Symptom & Missing ratio & Symptom & Missing ratio \\ \hline
difficulty sleeping &  0.0314 &sweats    &0.0275   &weight loss    &0.0236   \\
worrying   & 0.0362  &hot flashes    &0.0275   &increased appetite    &0.0259   \\
feeling sad  &0.0306   &sexual Interest   &0.0283   &itching    &0.0291   \\
feeling irritable   &0.0251   &short breath    &0.0228   &hair loss    &0.0330  \\
feeling nervous  &0.0346   &difficult breathing    &0.0251   &changes in skin    &0.0259 \\
concentrating  &0.0338  &cough    &0.0291   &like myself &0.0251   \\
energy lack  &0.0467   &chest tightness    &0.0220   &food tastes    &0.0236   \\
feeling drowsy   &0.0394   &weight gain    &0.0322  &lack of appetite    &0.0291   \\
mouth sores    &0.0212   &swallowing  &0.0181   &dry mouth    &0.0338   \\
vomiting    &0.0174   &nausea    &0.0370   &constipation    &0.0291   \\
diarrhea   &0.0236  &  cramps &0.0244 & bloated &0.0236 \\
swelling &0.0174  & pain &0.0378 & numbness &0.0402 \\
dizziness    &0.0197   & urination &0.0197 & &\\\hline
\end{tabular}}
\label{USCFMissing}
\end{table}


%\begin{figure}[h]
%\centering
%\includegraphics[width =0.9\linewidth]{images/PCA original .png}
%\caption{PCA comparison of the USCF augmented and original data}
%\label{PCA2}
%\end{figure}

%\begin{figure}[h]
%\centering
%\includegraphics[width =0.9\linewidth]{images/PCA augmented HMCM.png}
%5\caption{PCA comparison of the USCF augmented and original data}
%\label{PCA1}
%\end{figure}
